\documentclass[10pt,twocolumn,letterpaper]{article}

\usepackage{wacv}
\usepackage{times}
\usepackage{epsfig}
\usepackage{graphicx}
\usepackage{amsmath}
\usepackage{amssymb}
\usepackage{enumitem}
\usepackage{footmisc}

\usepackage{algorithm}
\usepackage{algorithmic}

\usepackage{amsmath}
\usepackage{amsfonts}
\usepackage{mathtools}

\usepackage{flushend}
\usepackage{float}

\usepackage{dblfloatfix}
\usepackage{verbatim}
\usepackage{pifont}
\newcommand{\cmark}{\ding{51}}%
\newcommand{\xmark}{\ding{53}}%

\def\wacvPaperID{0503} 

\wacvfinalcopy 

\ifwacvfinal
\def\assignedStartPage{1} 
\fi

\ifwacvfinal
\usepackage[breaklinks=true,bookmarks=false]{hyperref}
\else
\usepackage[pagebackref=true,breaklinks=true,colorlinks,bookmarks=false]{hyperref}
\fi

\ifwacvfinal
\else
\fi

\begin{document}

\title{Meta-UDA: Unsupervised Domain Adaptive Thermal \\  Object Detection using Meta-Learning }

\author{Vibashan VS$^{1}$\hspace{0.1mm}, Domenick Poster$^{2}$\hspace{0.1mm}, Suya You$^{3}$\hspace{0.1mm}, Shuowen Hu$^{3}$\hspace{0.1mm},\and Vishal M. Patel$^{1}$ \\
$^{1}$ Johns Hopkins University, MD, USA, 
$^{2}$ West Virginia University, WV, USA \and 
$^{3}$ U.S. Army CCDC Army Research Laboratory, USA \\
{\tt\small \{vvishnu2,vpatel36\}@jhu.edu, dposter@mix.wvu.edu, \{suya.you.civ,shuowen.hu.civ\}@mail.mil}
}

\maketitle

\begin{abstract}
 Object detectors trained on large-scale RGB datasets are being extensively employed in real-world applications. However, these RGB-trained models suffer a performance drop under adverse illumination and lighting conditions. Infrared (IR) cameras are robust under such conditions and can be helpful in real-world applications. Though thermal cameras are widely used for military applications and increasingly for commercial applications, there is a lack of robust algorithms to robustly exploit the thermal imagery due to the limited availability of labeled thermal data.  In this work, we aim to enhance the object detection performance in the thermal domain by leveraging the labeled visible domain data in an Unsupervised Domain Adaptation (UDA) setting. We propose an algorithm agnostic meta-learning framework to improve existing UDA methods instead of proposing a new UDA strategy. We achieve this by meta-learning the initial condition of the detector, which facilitates the adaptation process with fine updates without overfitting or getting stuck at local optima. However, meta-learning the initial condition for the detection scenario is computationally heavy due to long and intractable computation graphs. Therefore, we propose an online meta-learning paradigm which performs online updates resulting in a short and tractable computation graph. To this end, we demonstrate the superiority of our method over many baselines in the UDA setting, producing a state-of-the-art thermal detector for the KAIST and DSIAC datasets. 
 \vskip-3.5mm
\end{abstract}

\section{Introduction}

 \begin{figure}[t]
 \centering
\includegraphics[width=.9\linewidth]{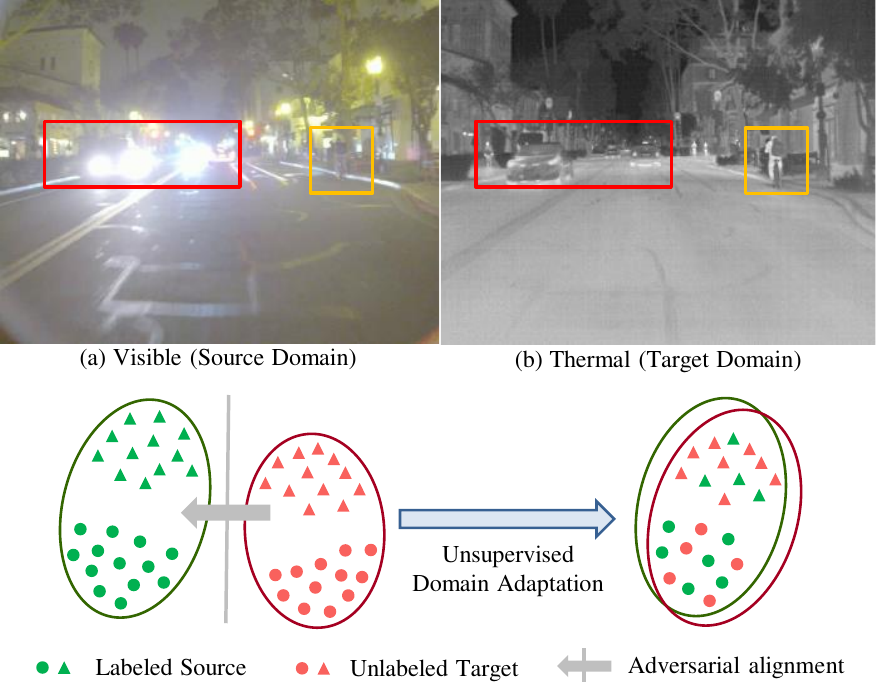}
\caption{In the top row, a comparison between the visible and thermal domains is illustrated. The red box corresponds to the region with adverse illumination and the yellow box corresponds to the region with low lighting conditions. The top row comparison shows that thermal images are more robust to adverse illumination and lighting conditions, resulting in better image representation than visible images. The bottom row shows the unsupervised domain adaptation setting, where a domain shift is mitigated between labeled visible domain and unlabeled thermal domain by performing adversarial feature alignment.}
\label{fig:motive} 
\vskip-4.5mm
\end{figure}

Object detection is a well-known problem in computer vision which has been actively researched for over two decades. With recent developments in deep Convolutional Neural Networks (CNNs) \cite{krizhevsky2012imagenet}, CNN-based object detectors produce state-of-the-art performance in many benchmark datasets. These advancements have enabled object detection as a fundamental component in perception systems for many real-world applications such as autonomous driving, surveillance and human activity recognition \cite{liu2020deep}. These object detection models are generally trained on large-scale RGB datasets such as ImageNet \cite{deng2009imagenet}, MS-COCO \cite{lin2014microsoft} and Pascal-VOC \cite{everingham2010pascal}. However, these RGB-trained models \cite{liu2016ssd, redmon2015you, ren2015faster} fail due to the domain shift under adverse illumination, occlusion, and lighting conditions. A recent study by NTSB \cite{NTSBstudy} showed that accidents caused by autonomous driving are due to a lack of sensory information regarding the surroundings and do not adequately detect pedestrians and vehicles. In addition, 75\% out of 5,987 U.S. pedestrian fatalities that happened in 2016 are during nighttime \cite{retting2017}. Hence, it is highly challenging for an autonomous system to decide solely based on visible sensory information, as visible sensors fail under such conditions (see Figure \ref{fig:motive}). 

In contrast, the Infrared (IR) sensors are robust under adverse illumination and  nighttime conditions and capture more information than visible sensors under those scenarios. Some of these thermal IR cameras are compact, low-cost and small in size. As a result, thermal IR cameras have become increasingly popular in many applications such as autonomous driving, drones and military surveillance for enhanced sensory information \cite{gade2014thermal}. Hence, addressing the detection and classification of objects in thermal imagery plays an essential role in deploying such models in the aforementioned real-world applications. Nevertheless, directly deploying the detectors trained on large-scale visible domain (RGB) datasets fail to detect objects in the thermal domain. This is due to the shift between the visible and thermal domains resulting in performance drop.  This is mainly due to the following reasons. First, the RGB-trained detectors lack generalizability and are not robust to unseen domains. Second, shortage of annotated large-scale datasets in the thermal domain. These two reasons hinder the performance level of the detectors from detecting and classifying objects in the thermal domain compared to the visible domain. 

Domain shift can be tackled by solving the lack of annotated data issues or increasing the model generalizing capacity and robustness. However, including more annotated data while training the detector is not feasible, as annotating data is a labor-intensive and time-consuming task. This leaves us with the only viable solution to improve the model's generalizing capability and make it robust by realizing the domain shift. Several domain adaptation settings \cite{chen2018domain, Saito2018StrongWeakDA} and methods have been proposed to mitigate the domain shift between the source (i.e. visible) and target (i.e. thermal) domain. In this work, we explore an unsupervised domain adaptation (UDA) setting \cite{ganin2015unsupervised}, where domain alignment is achieved solely by learning from the unlabeled target (thermal) domain. 

Most UDA works try to mitigate the domain shift using adversarial domain training \cite{chen2018domain, saito2019strong, vs2021mega}, pseudo-label self-training \cite{khodabandeh2019robust, roychowdhury2019automatic} or image-to-image translation techniques \cite{chen2020harmonizing, hsu2020progressive}. In this work, we tackle the domain shift problem by proposing an algorithm agnostic meta-learning strategy for domain adaptive detection instead of proposing a new DA strategy. The proposed meta-learning strategy is compatible with all existing UDA methods and it enhances their overall performance.  The performance improvement is possible because meta-learning learns the model learning process over multiple training iterations. As a result, meta-learning optimizes the adaptation process by achieving fine DA updates without overfitting or getting stuck at local optima. In meta-learning, there are two steps; 1)  base/inner learning - an inner learning algorithm that performs task-specific optimization. 2) meta/outer learning - an outer learning algorithm that improves base learning meta-parameters to satisfy the outer/meta objective. Thus, a meta-learning pipeline performs optimization at two levels to improve model learning, such as model initialization or model optimizer, to achieve meta-objectives such as learning speed or few-shot learning performance  \cite{krizhevsky2012imagenet, badrinarayanan2015segnet, ren2015faster}. As meta-objectives are more generic (i.e., model initialization, model optimizer), this can be extended to any existing algorithm resulting in improved performance. For the UDA detection, the meta-objectives are to minimize the supervised loss and reduce the domain shift source and target. However, performing meta-learning in an UDA detection setting is challenging for two reasons: i) object detection method such as Faster-RCNN is a computationally heavy framework and calculating meta-objectives for all training samples is intractable to perform meta updates. ii) Unlabeled target images provide no supervised loss to perform base/inner learning updates \cite{li2020online}. To overcome these challenges, we propose an online meta-learning paradigm where instead of performing meta-update after iterating over all the training samples, we perform intermittent meta-update during training. To demonstrate the effectiveness of the proposed method, we evaluate it on visible and thermal benchmark datasets and adaptation protocols and achieves state-of-the-art performance in all datasets. Moreover, ours is the first work to explore unsupervised domain adaptation for thermal object detection. 
Our main contributions are summarized as follows:
\begin{itemize}[noitemsep]
 	\item We introduce an algorithm agnostic meta-learning framework for thermal object detection in an unsupervised domain adaptation setting.
 	\item We propose an online meta-learning strategy which performs online meta-adaptation resulting in a short and tractable computation graph.
 	\item We empirically demonstrate the algorithm agnostic nature of our meta-learning framework over the existing domain adaptation algorithm and proposed architecture in the UDA setting, producing state-of-the-art performance on the KAIST and DSIAC datasets. 
 \end{itemize}


\section{Related work}
\noindent {\bf{Object detection.}}
Object detection is a fundamental problem being explored by the computer vision community for a long time due to its widespread real-world applications. Classical methods perform object detection based on object proposals obtained from selective search \cite{van2011segmentation}, super-pixel grouping \cite{levinshtein2010optimal} and HOG detector \cite{dalal2005histograms}.  The rise of deep CNNs shifted the object detection paradigm and resulted in state-of-the-art detectors. CNN-based detectors can be broadly classified into two categories i) One-stage detector and ii) Two-stage detector. One-stage detectors are YOLO \cite{redmon2015you} and SSD \cite{liu2016ssd}, whereas two-stage detectors are RCNN \cite{girshick2013rich}, Fast-RCNN \cite{girshick2015fast}, Faster-RCNN \cite{ren2015faster}. One-stage detectors perform object classification and bounding box regression in a single pipeline. In contrast, the two-stage detectors perform object detection at two stages. In the first stage, a region proposal network is used to generate object proposals and in the second stage,  object proposal undergoes object classification and bounding box regression.  However, all of these state-of-the-art detectors' performance drops under domain shift.\\
\noindent {\bf{Thermal object detection.}}
Thermal object detection plays a vital role in detecting objects in surveillance and military operation \cite{gade2014thermal}. In \cite{9342179}, the SSD architecture is used to detect objects in the thermal domain. \cite{9133581} proposed to use the YOLO architecture to detect objects in the thermal domain. Dai \etal \cite{dai2021tirnet} proposed TIRNet to detect objects in thermal IR images to provide more sensory information for autonomous driving. In order to exploit both visible and thermal domains, Devaguptapu \etal \cite{devaguptapu2019borrow} proposed a detection framework where they fuse visible and thermal features at a high level to capture more information resulting in better detection. Later in \cite{munir2021sstn}, they propose a self-training method to enhance the performance in the thermal domain using both visible and thermal images.  Note that all of these works have neglected to address a more practical scenario where we have access to a large-scale labeled visible domain image dataset and adapt the detector to the unlabeled thermal domain images.\\ 
\noindent {\bf{Unsupervised domain adaptive object detection.}}
In object detection,  Chen \etal \cite{Chen2018DomainAF} was the first to explore unsupervised domain adaptation settings. In particular, Chen \etal \cite{Chen2018DomainAF} proposed DA Faster-RCNN network, which performs adversarial domain training to mitigate the domain shift at the image and instance levels. Later, Saito \etal \cite{Saito2018StrongWeakDA} noted that weak alignment of the global features and strong alignment of the local features plays a significant role in adaptation.  Cai \etal \cite{cai2019exploring} performed domain adaptive detection using a mean-teacher framework to utilize the unlabeled target data better.  Recently, Sindagi \etal \cite{sindagi2020prior} proposed the use of weather priors for adapting detectors to different weather conditions. Zhu \etal \cite{zhu2019adapting} performed a region mining strategy in order to perform a region-level alignment and showed its benefits compared to conventional domain adversarial training. In addition, there are many other works that have addressed domain adaptive object detection in 2D \cite{oza2021unsupervised, roychowdhury2019automatic, vs2021mega} and 3D \cite{saltori2020sf, hegde2021uncertainty} domains. However, no works have explored unsupervised domain adaptation settings for thermal object detection. In this paper, we investigate unsupervised domain adaptation for thermal object detection.

\noindent {\bf{Meta-learning.}}
In conventional deep learning, for a given task, models are optimized according to task-specific loss resulting in minimum prediction error \cite{krizhevsky2012imagenet, badrinarayanan2015segnet, ren2015faster}. However, meta-learning provides an alternative paradigm where the model learns to learn over multiple training episodes \cite{thrun1998learning}. In other words, meta-learning is a process of learning to the learn algorithm over multiple training episodes. The meta-learning landscape can be divided into three parts -- meta-optimizer, meta-representation, and meta-objective. Meta-optimizer is the choice of optimizer used to learn how the optimization works in the outer loop of meta-learning \cite{finn2017model, houthooft2018evolved}.  Meta-representation aims what meta-knowledge is to be learned and updated in the process of meta-learning \cite{finn2017model}. Finally, the meta-objective is the goal of the meta-learning task to be achieved at the end of training \cite{li2019feature, finn2017model, mishra2017simple}. Therefore in this work, we investigate the meta-learning framework for the UDA detection setting, where the meta-representation is the initial condition of the detector and the meta-objective is the detection and adaptation losses.

 \begin{figure*}[h]
 \centering
\includegraphics[width=1.0\linewidth]{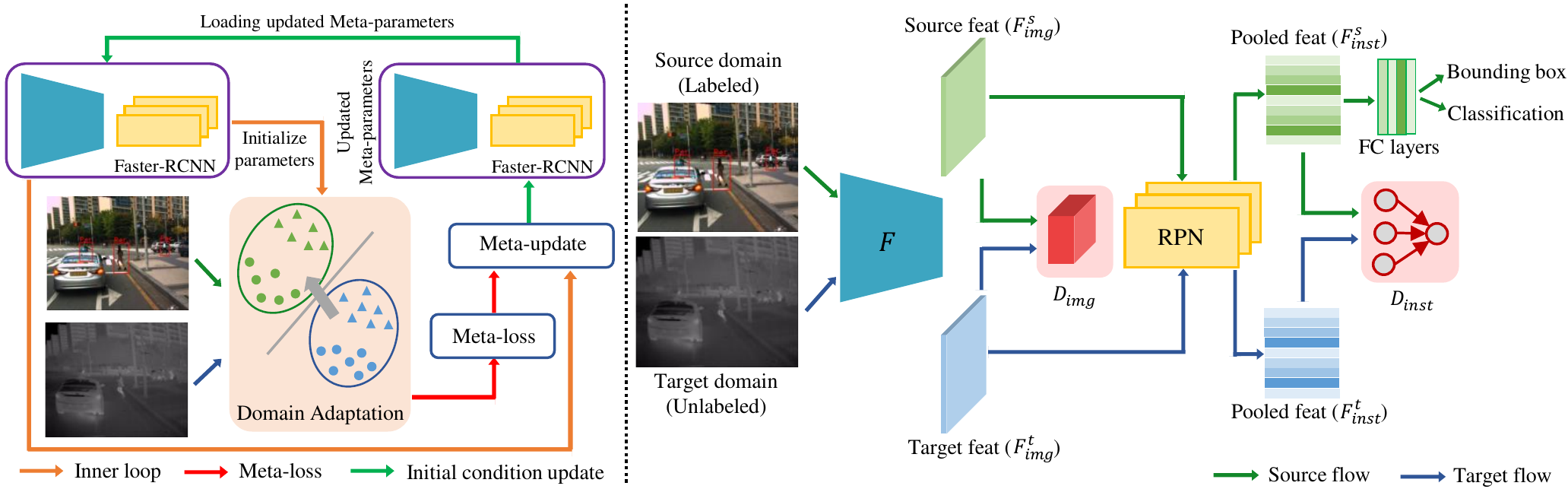}
\caption{Overview of meta-learning strategy is presented on the left side and adaptation framework is presented on the right side. The meta-learning performs bi-level optimization, where the inner loop performs domain adaptation and the outer loop performs meta-update with respect to meta-loss. The updated meta-parameters are loaded as the initial condition of Faster-RCNN and this process is repeated. The adaptation framework performs adversarial feature alignment between source and target domain at the image and instance level.}
\label{fig:archi} 
\vskip-3.5mm
\end{figure*}

\section{Proposed method}\label{sec:proposed_method}

\subsection{Preliminaries}
Conventional unsupervised domain adaptation methods assume that both source and target data are available while adapting a model for the target domain. Formally in the UDA setting,  we denote the labeled source domain as ${D}_s = \{X_s^n, y_s^n\}_{n=1}^{N_s}$ where $X_s^n$ denotes the $n^{th}$ source image and $y_s^n$ denotes the corresponding object category and bounding box ground truth. In the target domain, the unlabeled target images are denoted as ${D}_t = \{X_t^n\}_{n=1}^{N_t}$ where $X_t^n$ denotes the $n^{th}$ unlabelled target image. Following the standard domain adaptive detection works, we employ Faster-RCNN ($\Theta$) \cite{ren2015faster} with the VGG16 \cite{simonyan2014very} backbone as our detector. Unsupervised domain adaptive detection aims to train a detector on the labeled source images and exploit unlabelled target images to reduce the domain shift. To achieve this, we propose an algorithm agnostic meta-learning framework for the thermal object detector that optimizes the model initial condition for domain adaptive detection.

\subsection{Detection framework}
The Faster-RCNN pipeline consists of three main components: an encoder, a Region Proposal Network (RPN) and a region classification and regression network (RCNN). For a given image $X^{n}$, the encoder extracts the feature map and then on top of the extracted feature map, RPN generates class agnostic object region proposals. Following that, the RCNN network classifies object category and predict bounding box offset for the object proposal. The training objective of the detector pipeline is to minimize the RPN and RCNN loss as follows:
\begin{equation}
\mathcal{L}_{det}(X^{n}, Y^{n}) = \mathcal{L}_{cls}^{rpn} +\mathcal{L}_{reg}^{rpn} +\mathcal{L}_{cls}^{rcnn}+\mathcal{L}_{reg}^{rcnn}.\\
\end{equation}

where $L_{cls}^{rpn}$ and $L_{cls}^{rcnn}$ are the object classification loss \cite{ren2015faster} and $L_{reg}^{rpn}$ and $L_{reg}^{rcnn}$ are the bounding box regression loss \cite{ren2015faster} for RPN and RCNN network.

\subsection{Image and instance level adaptation}
Faster-RCNN \cite{ren2015faster} is a two-stage detector that performs detection at two levels. When a source trained Faster-RCNN encounters images from the target domain, the performance drops due to domain shift, affecting the detector at two levels. These two levels of the detector are image level and instance level. Image level represents the encoder feature output and instance-level represents the RPN feature output. To mitigate the domain shift, we employ an adversarial domain classifier at both image and instance levels. The adversarial domain classifier helps to align distribution shift, resulting in domain invariant features at the image and instance levels. Briefly, performing adversarial alignment at the image level ensures global feature alignment, such as the shift in image style, illumination. Performing adversarial alignment at the instance level ensures local feature alignment, such as the shift in object size, style, viewpoint, etc. In our work, we have extended the discriminator architecture proposed in DA Faster-RCNN \cite{Chen2018DomainAF} to obtain a stronger and robust classifier which helps in better feature alignment. Architecture details are presented in the supplementary material. 

First, let us denote the image-level domain classifier as $\mathcal{D}_{img}$ which classifies the input encoded features as  source or target domain. For given source ($X_s^n$) and target ($X_t^n$) domain images, the encoder extracted feature map are denoted as $F_{img}^s, \ F_{img}^t \in \mathbb{R}^{C \times H \times W}$. Feeding $F_{img}^s, \ F_{img}^t$ to  $\mathcal{D}_{img}$ outputs a prediction map of size $H \times W$ with domain labels are set to 1 and 0 for source and target domain respectively. The  least squared loss is used to supervise the domain classifier with domain label $y_d \in {0, 1}$ and the loss function can be written as:

\begin{multline}
\mathcal{L}_{img}(X_s^{n}, X_t^{n}) = -\sum_{h=1}^{H} \sum_{w=1}^{W} y_d (1-\mathcal{D}_{img}({F_{img}^{s}}^{(h,w)}))^{2} \\ 
+(1-y_d) (\mathcal{D}_{img}({F_{img}^{t}}^{(h,w)}))^{2}.
\label{eq:img_align}
\end{multline}

Second, let us denote instance-level domain classifier as $\mathcal{D}_{inst}$ which classifies the RPN pooled features as source or target domain. For given source ($X_s^n$) and target ($X_t^n$) domain images, the RPN pooled features are denoted as $F_{inst}^s, \ F_{inst}^t \in \mathbb{R}^{C \times D}$. We feed $F_{inst}^s, \ F_{inst}^t$ to  $\mathcal{D}_{inst}$ which outputs a prediction map of size $D$ with domain labels  set to 1 and 0 for source and target domain, respectively. The  least squared loss is used to supervise the domain classifier and the loss function can be written as:
\begin{multline}
\mathcal{L}_{inst}(X_s^n, X_t^n) = -\sum_{d=1}^{D} y_d (1-\mathcal{D}_{inst}({F_{inst}^s}^{(h,w)}))^2 \\
+(1-y_d) (\mathcal{D}_{inst}({F_{inst}^t}^{(h,w)}))^2.
\label{eq:inst_align}
\end{multline}
To achieve the domain alignment, we utilize the Gradient Reversal Layer (GRL)  \cite{ganin2014unsupervised}, which flips the gradient sign after propagating gradient through the domain classifier. Therefore, when minimizing \ref{eq:img_align} and \ref{eq:inst_align} for image and instance-level domain classifiers, the GRL helps in achieving equilibrium. In the equilibrium condition, the input features are domain invariant and the domain classifier cannot differentiate the source and target features. Furthermore, we opt for least-squares loss instead of using binary-cross entropy loss, as it is shown to be working better \cite{mao2017least}.  Hence, the total domain adaptation loss is formulated as follows
\begin{equation}
\mathcal{L}_{da}(D_s,D_t) = \mathcal{L}_{img}(D_s,D_t) +\mathcal{L}_{inst}(D_s,D_t).
\end{equation}
 \begin{figure}[t]
    \includegraphics[width=0.9\linewidth]{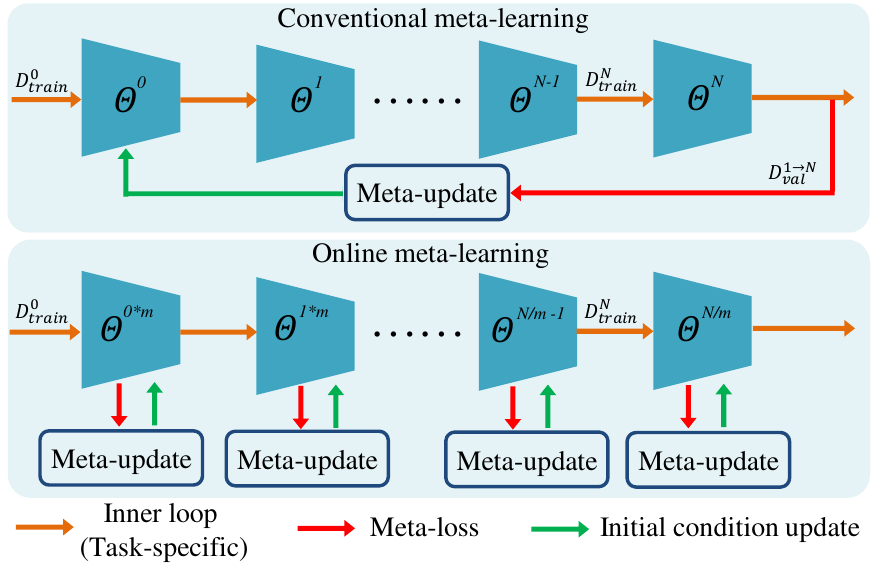}
    \caption{In conventional meta-learning (top-row), the meta-loss is computed for the model obtained from the inner loop after iterating over the complete train set. As a result, this leads to a long and intractable computation graph. In the proposed method, we compute meta-loss at regular interval $m$ during training resulting in a short and tractable computation graph.}
\label{fig:online} 
\vskip-5.5mm
\end{figure}

\subsection{Overall training objective}
In the UDA setting, we have labeled samples from the source domain $D_s$ and unlabeled samples from the target domain $D_t$. The objective of the detector is to learn from labeled source data by minimizing supervised loss ($L_{det}$). In addition, for domain adaptation, the detector should be domain invariant and this can be realized by reducing the domain shift by minimizing the adaptation loss $L_{da}$ obtained from source data and target data. Hence, the overall loss for unsupervised domain adaptation setting is defined as:
\setlength{\belowdisplayskip}{0pt} \setlength{\belowdisplayshortskip}{0pt}
\setlength{\abovedisplayskip}{0pt} \setlength{\abovedisplayshortskip}{0pt}
\begin{equation}
\mathcal{L}_{uda}(D_s,D_t) = \mathcal{L}_{det}(D_s) + \mathcal{L}_{da}(D_s,D_t).
\label{eq:uda}
\end{equation}

The degree of domain alignment depends on the model optimization strategy. Thus, meta-learning the detector's initial condition helps in achieving fine DA updates, resulting in an optimally adapted detector.

\subsection{Online meta-adaption}
Conventional meta-learning \cite{finn2017model} the initial condition can be expressed as a bi-level optimization problem, where the inner loop optimizes according to the task-specific loss and the outer algorithm optimizes the meta-parameters with respect to meta-loss as shown in Figure \ref{fig:archi}. Thus, meta-learning the initial condition is formulated as:
\begin{equation}\label{eq:meta}
\begin{aligned}
    \Theta & =\underbrace{\underset{\Theta}{\operatorname{argmin} } ~~ \mathcal{L}_{\text{outer}}( \overbrace{\mathcal{L}_{\text{inner}}(\Theta, \mathcal{D}_{\text{tr}})}^{\text{Inner-level}}, \mathcal{D}_{\text{val}})}_{\text{Outer-level}}, 
\end{aligned}
\end{equation}
where $\Theta$ is the meta-parameters which is initialized model parameters; $D_{tr}$ and $D_{val}$ are train and validation dataset; $L_{inner}$ denotes the task-specific loss on its training set and the $L_{outer}$ denotes validation loss obtained after inner optimization. The overall objective of Eqn \eqref{eq:meta} is to obtain an optimal initial condition such that the validation loss is minimum. Extending this to UDA setting, we create a train and validation dataset for source and target domain denoted as ${D}_s^{tr}$, ${D}_t^{tr}$ and ${D}_s^{val}$ , ${D}_t^{val}$ respectively from ${D}_s$ and ${D}_t$. In the inner loop, the task-specific loss corresponds to $L_{uda}$ obtained from the train set ${D}_s^{tr}$ and ${D}_t^{tr}$ and is computed as follows:
\begin{equation}
 \Theta_n^{'} = \Theta - \alpha  \nabla_\Theta \mathcal{L}_{uda}(D_s^{tr}(n),D_t^{tr}(n)),
\end{equation}
where $n$ corresponds to the $n^{th}$ sample from source and target training set and $\alpha$ is the inner loop learning rate. In the outer loop, the meta-loss is computed on the validation set for the inner loop model, which is obtained after fully iterating over the training set. Following that, the initial condition of the detector (i.e. meta-parameters) are updated with respect to meta loss as follows
\begin{algorithm}[t]
\caption{Online meta-adaptation for UDA}
\label{alg:mamlsup}
\begin{algorithmic}[1]
{\footnotesize
\REQUIRE ${D}_s^{tr}$, ${D}_t^{tr}$, ${D}_s^{val}$, ${D}_t^{val}$
\REQUIRE $\alpha$, $\beta$: meta learing rate hyperparameters
\STATE randomly initialize $\Theta$
\WHILE{not done}
  \FOR{$m$}
      \STATE Sample batch of ${D}_s^{tr}$, ${D}_t^{tr}$, ${D}_s^{val}$, ${D}_t^{val}$
      \STATE Evaluate $\mathcal{L}_{uda}(D_s^{tr}(n),D_t^{tr}(n))$ using Equation~(\ref{eq:uda})
      \STATE Compute adapted parameters with gradient descent: 
      \STATE $\Theta_i'=\Theta-\alpha \nabla_\Theta  \mathcal{L}_{uda}$
      \STATE Compute  Meta-loss for $\Theta_i'$ using Equation~(\ref{eq:uda}): 
      \STATE $\mathcal{L}_{uda}^{meta loss}$ += $\mathcal{L}_{uda}(D_s^{val},D_t^{val})$
 \ENDFOR
 \STATE Update $\Theta \leftarrow \Theta - \beta \nabla_\Theta \mathcal{L}_{uda}^{meta loss}$
\ENDWHILE
}
\end{algorithmic}
\label{algo:meta}
\end{algorithm}
\begin{equation}
 \Theta = \Theta - \beta \nabla_\Theta \overbrace{\sum_{n=1}^{N} {L}_{uda}(D_s^{val},D_t^{val})}^{\text{Meta-loss}},
 \label{eq:meta_update}
\end{equation}
where $\beta$ is the meta-learning rate. Thus, we learn to learn the optimization process, resulting in fine DA updates without overfitting or getting stuck at local optima \cite{finn2017model, li2020online}. However, meta-learning is not compatible with the domain adaptive detection framework. Because storing all the inner-loop computation graphs in the detection pipeline is computationally heavy and backpropagating through them is intractable. Thus, we propose an online meta-domain adaptation strategy for the detection pipeline, which performs online meta-adaptation resulting in a short and tractable computation graph. In other words, we extended the meta-learning paradigm to perform on-the-fly meta-updates by optimizing inner and outer loops for intermittent steps as shown in Figure \ref{fig:online}. As per Algorithm \ref{algo:meta}, we perform online meta-adaptation for UDA setting by alternatively optimizing inner and outer loops at short intervals $m$.  This avoids the long computational graphs and provides stable optimization for DA updates. Moreover, the on-the-fly paradigm understands a better association between the initial condition and meta-loss compared to conventional meta-learning. Thus, the online meta-learning ensures gradual optimization and achieves proper fine-tuning for the initial condition resulting in an enhanced adapted detector with more robustness and generalizing capability. \vskip-2.5mm

\vskip-2.5mm
\section{Experiments and results}
In this section, we evaluate the proposed method to empirically show its effectiveness on two different adaptation scenarios with visible to thermal domain shift experiments: 1. Automatic Target Recognition \cite{dsiac}, and 2. Multi-spectral Pedestrian Detection \cite{hwang2015multispectral}. 


 \begin{figure}[h]
 \centering
\includegraphics[width=.9\linewidth]{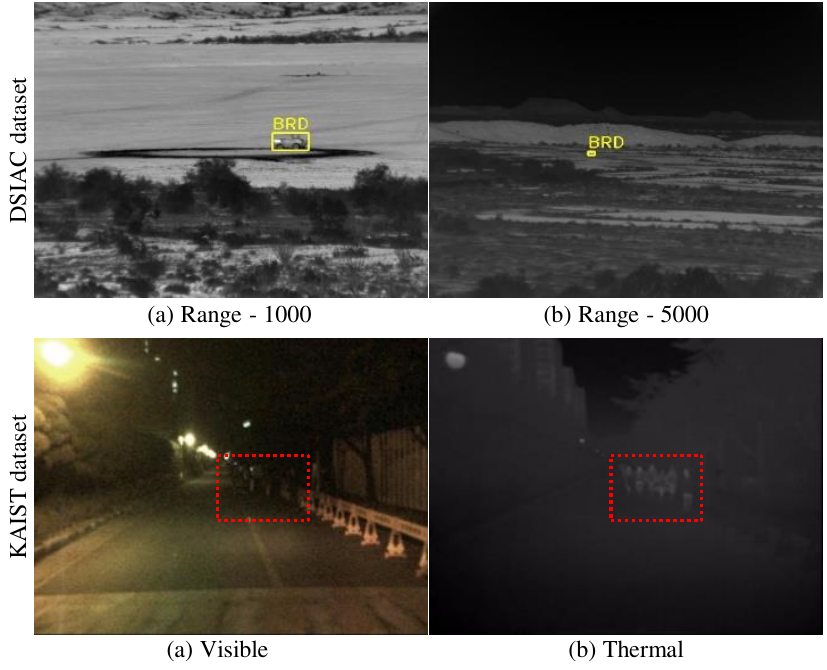}
\caption{Sample images from the DSIAC thermal dataset at ranges 1000 and 5000 are shown on the top row.  We can observe that the object at Range 5000 is very small compared to range 1000 and is not even visible to the naked eyes. However, detecting these objects is crucial for military surveillance. Sample images from the KAIST dataset are shown in the bottom row where the red box denotes the region with pedestrians. We can observe that thermal images can capture more salient features than visible images during nighttime resulting in better image representation.}
\label{fig:data_sample} 
\vskip-4.5mm
\end{figure}

\noindent {\bf{Automatic target recognition.}}
Automatic Target Recognition (ATR) is an algorithm's ability to perform real-time target recognition using multiple sensory information. ATR is a well-known problem setup and has been employed in unmanned systems in automated surveillance and military operations \cite{dsiac}.  Because these unmanned aerial vehicles (UAVs) and unmanned ground vehicles (UGVs) have multiple sensors to capture different data modes facilitating ATR algorithms. These data are from the visible and thermal domain incorporated to achieve an accurate and robust ATR system. However, most publicly available datasets have labeled visible images and lack labeled samples for thermal images.  Hence, it is important to train the detector for the thermal domain to enable ATR in surveillance and military operations. 

We implement our method for the ATR problem setting and evaluate our method on the publicly available DSIAC dataset provided by the US Army Night Vision and Electronic Sensors Directorate (NVESD) \cite{dsiac}. The DSIAC dataset contains 106 GB of visible data and 207 GB of Middle Wavelength Infrared (MWIR) data. It contains eight classes, among which two are civilian vehicles and remaining six are  military vehicles : `Pickup', `Sport vehicle', `BTR70', `BRDM2', `BMP2', `T72', `ZSU23', `2S3'. This dataset was collected during day and night time, with each video sequence containing 1800 video frames. The distance between cameras and targets are varied from 500 to 5000 meters at an interval of  500 meters. Following the conventional UDA setting, we assume we have access to labeled visible data and unlabeled thermal data. For both domains, we sample every ten frames from the dataset video sequence for the ranges 1000-5000 at interval 1000. Further, we assign 75 \% of the sampled data as the training set and 25 \% of the sampled data as the test set for each range. Thus in this work, we investigate the domain shift between visible and thermal domains at different ranges for the ATR problem.

\begin{table*}[t]
\caption{Quantitative results (mAP) for visible $\rightarrow$ thermal adaptation for the DSIAC dataset. Source only: Trained on visible domain and tested on thermal domain; Oracle: Trained on visible thermal and tested on thermal domain}
\centering
\label{tab:atr}
\resizebox{0.9\linewidth}{!}{\begin{tabular}{|c|ccc|cccccccc|c|}
\hline
{Method} & \multicolumn{3}{c|}{Range}           & {Pkup Trck} & {Sprt Vech} & {BTR70} & {BRDM2} & {BMP2} & {T72} & {ZSU23} & {2S3} & {mAP} \\ \cline{2-4}
                        & 1000       & 5000       & All        &                            &                            &                        &                        &                       &                      &                        &                      &                      \\ \hline \hline
Source Only             & \cmark & \xmark & \xmark & 26.9                       & 20.4                       & 81.6                   & 23.9                   & 28.7                  & 46.3                 & 6.0                    & 29.4                 & 32.2                 \\ 
Source Only             & \xmark & \cmark & \xmark & 0.0                        & 0.0                        & 0.0                    & 0.0                    & 0.0                   & 0.0                  & 0.0                    & 0.0                  & 0.0                  \\ 
Source Only             & \xmark & \xmark & \cmark & 3.3                        & 4.5                        & 9.5                    & 10.3                   & 0.7                   & 13.3                 & 0.2                    & 5.0                  & 5.2                  \\ \hline
DA-Faster               & \cmark & \xmark & \xmark & 9.4                        & \textbf{30.0}                       & 79.6                   & 38.9                   & 37.8                  & 44.5                 & \textbf{3.6}                    & 44.8                 & 36.1                 \\ 
Ours                    & \cmark & \xmark & \xmark & \textbf{16.6}                       & 21.2                       & \textbf{81.0}                   & \textbf{67.4}                   & \textbf{41.1}                  & \textbf{51.5}                 & 1.5                    & \textbf{55.5}                 & \textbf{41.9}                 \\ \hline
DA-Faster               & \xmark & \cmark & \xmark & 0.0                        & 0.0                        & 0.0                    & 0.0                    & 0.0                   & 0.0                  & 0.0                    & 0.0                  & 0.0                  \\ 
Ours                    & \xmark & \cmark & \xmark & 0.0                        & 0.0                        & \textbf{3.0}                    & 0.0                    & 0.0                   & 0.0                  & 0.0                    & \textbf{6.1 }                 & \textbf{1.1 }                 \\ \hline
DA-Faster               & \xmark & \xmark & \cmark & 4.2       & 11.7      & 12.1  & 11.4  & 0.3  & 9.1  & \textbf{0.2}   & 0.3  & 6.2                  \\ 
Ours                    & \xmark & \xmark & \cmark & \textbf{9.5}                        & \textbf{14.0}                       & \textbf{16.6}                   & \textbf{12.7}                   & \textbf{1.4}                   & \textbf{16.4 }                & \textbf{0.2}                    & \textbf{2.0}                  & \textbf{9.1}                  \\ \hline
Oracle                  & \cmark & \xmark & \xmark & 100                        & 100                        & 100                    & 100                    & 100                   & 100                  & 100                    & 100                  & 100                  \\ 
Oracle                  & \xmark & \cmark & \xmark & 21.1                       & 21.8                       & 35.8                   & 21.0                   & 30.0                  & 37.9                 & 1.0                    & 40.1                 & 26.1                 \\ 
Oracle                  & \xmark & \xmark & \cmark & 63.0                       & 66.3                       & 77.1                   & 68.9                   & 66.8                  & 69.1                 & 80.2                   & 79.9                 & 71.4                 \\ \hline
\end{tabular}}
\vskip-4.0mm
\end{table*}

\noindent {\bf{Multi-spectral pedestrian detection.}}
The KAIST Multi-Spectral dataset \cite{hwang2015multispectral} contains 95,000 8-bit  paired thermal and visible images. This dataset is collected during day and night using a FLIR A35 microbolometer LWIR camera with 320 $\times$ 256 pixels resolution. It contains a standard train-test split of 76000 train images and 19000 test images with only annotation available for the pedestrian class. Following the conventional UDA setting, we assume we have access to labeled visible and unlabeled thermal data and adapt the detector to the thermal domain. 

\begin{table}[h]
\caption{Quantitative results (mAP) for visible $\rightarrow$ thermal adaptation on the KASIT dataset.}
\centering
\label{tab:kaist}
\begin{tabular}{|c|c|c|c|}
\hline
Method      & Meta-learn  & Person & mAP  \\ \hline \hline
Source Only & \xmark  & 9.1    & 9.1  \\ \hline
DA-Faster   & \xmark  & 11.9   & 11.9 \\ 
Ours        & \xmark  & 23.2   & 23.2 \\ \hline
DA-Faster   & \cmark &  13.7   & 13.7    \\ 
Ours        & \cmark &  \textbf{24.6}   &  \textbf{24.6}    \\ \hline
Oracle      & \xmark  & 43.9   & 43.9 \\ \hline
\end{tabular}
\vskip-4.0mm
\end{table}

\noindent {\bf{Implementation details.}}
We adopt the unsupervised domain adaptation setting for all our experiments, where we have access to labeled visible data and unlabeled thermal target data.  By default, our base detector is Faster-RCNN and is initialized with pre-trained VGG weights.  The shorter side of the input images are resized to 600 pixels while maintaining the aspect ratio.  We perform random horizontal flip and subtract the image mean for all input images as part of data augmentation. During training, the batch size is set equal to 1. We set the domain adaptation loss weight  $\lambda$ equal to 0.1. The learning rate hyperparameter $\alpha$ and $\beta$ for the inner and outer loop meta-learning are set equal to 0.001.   Momentum is set equal to 0.9 for five epochs and then decreases the learning rate to 0.0001. In meta-learning, alternatively optimizing inner and outer loops for a short interval $m$ is set equal to 3. We train the network for ten epochs and use the mean average precision (mAP) metric as an evaluation metric. 

\subsection{Quantitative comparison}
We compare our method with the existing baselines and show our proposed method's effectiveness under different domain shift settings. Note that to the best of our knowledge, this is the first work addressing unsupervised domain adaptation for thermal detectors.   

\noindent \textbf{Automatic target recognition.} 
In Table \ref{tab:atr}, we report the performance of our method against DA Faster-RCNN baseline for different ranges. Range - 1000 and 5000 represent the distance at which the target has been captured with respect to the camera. Range - ``All" includes the range from 1000-5000 at intervals of 1000 meters. In addition, the oracle experiment denotes training and testing on the visible domain, whereas the source-only experiments indicate training on the visible domain and testing on the thermal domain. From Figure \ref{fig:data_sample} for Range-1000, we can observe that the targets are big and easy to distinguish. As a result, we obtain oracle performance as 100 mAP. However, visible to thermal domain shift affects the detector by a large margin in the source-only experiment as shown in Table \ref{tab:atr}.  From Figure \ref{fig:data_sample} for Range-1000, we can observe that the targets are very small and difficult to distinguish. Therefore, the oracle performance for Range - 5000 is only 26.8 and the corresponding source-only performance is 0 mAP. For Range -``All", the oracle and source-only performances are 71.4 mAP and 5.2 mAP, respectively. As can be seen from Table \ref{tab:atr}, domain shift causes catastrophic performance degradation. Hence, we argue that mitigating the domain shift effect plays a crucial role in deploying thermal detectors in real-world applications. Our meta-learning strategy  minimizes the domain shift by learning optimal DA updates for adaptation. 

As shown in Table \ref{tab:atr}, for Range-1000, our proposed method performs better than DA-Faster RCNN by 6.2 mAP. For Range-5000, even though after adaptation DA Faster-RCNN performance was 0 mAP, our proposed adaptation strategy ensures the optima adaptation updates, resulting in 1.1 mAP. For Range - ``All", our model achieves 40\% better mAP than the DA Faster-RCNN method. Thus, we empirically demonstrate that meta-learning the initial condition of the detector is improving the adapted detector's performance.
\begin{table*}[t]
\caption{Ablation study for meta-learning strategy in DSIAC dataset Range - ``All" for visible $\rightarrow$ thermal adaptation.}
\centering
\label{tab:ablation}
\resizebox{0.9\linewidth}{!}{\begin{tabular}{|c|c|cccccccc|c|}
\hline
{Method} & Meta-learn  & Pkup Trck & Sprt Vech & {BTR70} & {BRDM2} & {BMP2} & {T72}  & {ZSU23} & {2S3}  & mAP  \\ \hline \hline
Source Only                                           & \xmark  & 3.7       & 5.8       & 11.5  & 10.1  & \textbf{1.6}  & 9.1  & 0.2   & \textbf{2.8}  & 5.6  \\ \hline
DA-Faster                                             & \xmark  & 4.2       & 11.7      & 12.1  & 11.4  & 0.3  & 9.1  & 0.2   & 0.3  & 6.2  \\ 
Ours                                                  & \xmark  & \textbf{10.4}      & 12.9      & 12.1  & \textbf{12.9}  & 1.2  & 14.4 & \textbf{0.4}   & 1.5  & 8.2  \\ \hline
DA-Faster                                             & \cmark &   2.7        &  10.4         & \textbf{21.4}      &    9.4   &   1.0   &  12.0    &  \textbf{0.4}     &   0.1   &   7.1   \\ 
Ours                                                  & \cmark & 9.5       & \textbf{14.0}      & 16.6  & 12.7  & 1.4  & \textbf{16.4} & 0.2   & 2.0  & \textbf{9.1}  \\ \hline
Oracle                                                & \xmark  & 63.0      & 66.3      & 77.1  & 68.9  & 66.8 & 69.1 & 80.2  & 79.9 & 71.4 \\ \hline
\end{tabular}}
\vskip-2.0mm
\end{table*}

 \begin{figure*}[t]
 \centering
\includegraphics[width=.88\linewidth]{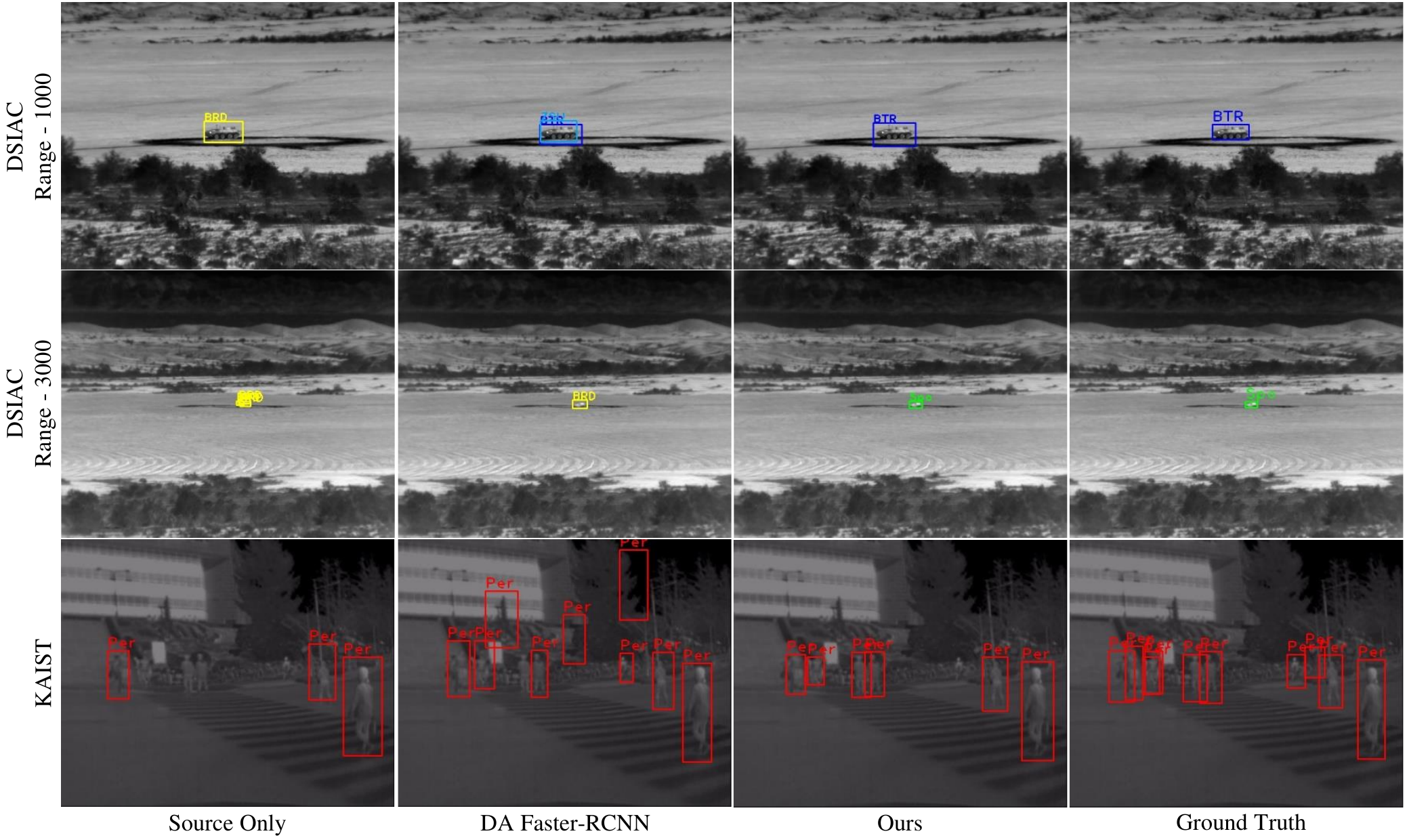}
\caption{Qualitative detection results. In the DSIAC dataset, source only and DA Faster-RCNN produce false-positive predictions, whereas our method recognizes the object correctly. Similarly, in the KAIST dataset, our method reduces false-positive as well as produces a high-quality prediction. Because meta-learning helps in achieving fine DA updates resulting in a more robust and generalized detector.}
\label{fig:quali} 
\vskip-4.0mm
\end{figure*}
\noindent \textbf{Multi-spectral pedestrian detection.} 
In the KAIST dataset, we perform pedestrian detection using Faster-RCNN which produces 43.9 mAP as the oracle performance. In the source-only experiment, the performance drops by 34.8 mAP due to the domain shift as shown in Table \ref{tab:kaist}.  DA Faster-RCNN performs adversarial feature alignment to mitigate the domain shift resulting in 11.9 mAP with an improvement of 2.8 mAP from the  source-only performance. Following our method, we obtain 21.7 mAP with an improvement of 41.1 mAP from the source-only performance. This difference in improvement shows the effectiveness of a strong discriminator even though our method is an extension of the DA Faster-RCNN approach. However, by applying meta-learning, the initial condition-based adaption ensures optimal adaption, resulting in improved performance for DA Faster-RCNN and our method by  13.7 mAP and 24.6 mAP, respectively.\\
\noindent {\bf{Ablation study.}}
We study the adaptation impact of our proposed meta-learning the initial condition strategy for DA Faster-RCNN and our framework. Table \ref{tab:ablation} presents the ablation analysis done on the DSIAC dataset for all ranges.  From Table \ref{tab:ablation}, we can infer that DA Faster-RCNN and our framework without meta-learning produce 6.2 mAP and 8.2 mAP, respectively. The improvement of our framework over DA Faster-RCNN is due to the utilization of a stronger discriminator resulting in a better feature alignment. Furthermore, employing the meta-learning the initial condition strategy for DA Faster-RCNN and our framework, we obtain  14.5 \% and 10.9 \% improvement for DA Faster-RCNN and our framework, respectively. This performance improvement using meta-learning supports our argument that meta-learning helps in learning to learn the adaptation process by updating the detector's initial condition. \\
\noindent {\bf{Qualitative comparison.}}
We visualize the detection performance of source only model, DA Faster-RCNN, our method with respect to the ground truth in Figure \ref{fig:quali}. The visualization is presented for experiments on the DSIAC dataset Range-1000 and All, KAIST dataset in first, second and third row, respectively as shown in Figure \ref{fig:quali}. In the first and second row, we can observe that the source-only model produces false positive detection due to the domain shift. Further, due to adaptation, DA Faster-RCNN recognizes the object correctly but with a few false-positive predictions. Similarly, for the KAIST dataset (third row), our method produces less miss detection compared to DA Faster-RCNN. Thus, our quantitative and qualitative analysis shows the effectiveness of the proposed method on the DSIAC and KAIST datasets. \vskip-2.5mm
\vskip-2.5mm
\section{Conclusion}
We presented an unsupervised domain adaptive thermal object detection framework for real-world applications. Specifically, we introduced an algorithm agnostic meta-learning framework applicable for existing domain adaptation techniques. Furthermore, we proposed an online meta-domain adaptation compatible with the detection framework, which performs online meta-adaptation resulting in a short and tractable computation graph. Finally, we demonstrated the algorithm agnostic nature of our meta-learning framework over the existing domain adaptation algorithm and proposed architecture in the UDA setting.  Our method produces state-of-the-art thermal detection performance on the KAIST and DSIAC datasets. \vskip-2.5mm
\vskip-2.5mm
\section{Acknowledgment}
Research was sponsored by the Army Research Office and was accomplished under Cooperative Agreement Number W911NF-20-2-0224. The views and conclusions contained in this document are those of the authors and should not be interpreted as representing the official policies, either expressed or implied, of the Army Research Office or the U.S. Government. The U.S. Government is authorized to reproduce and distribute reprints for Government purposes notwithstanding any copyright notation herein.

{\small
\bibliographystyle{ieee_fullname}
\bibliography{egbib}
}


\newpage
\clearpage

\appendix
\onecolumn

\begin{center}
  \textbf{
    \Large Supplementary Material for ''Meta-UDA: Unsupervised Domain \\ Adaptive Thermal  Object Detection using Meta-Learning''} \\
  \Large 
-- \\
 \textbf{\large Supplementary Material} \\
\hspace{1cm}
\end{center}


\section{Network architecture}
In Table~\ref{tab:img_da}, we show the architecture details for image-level domain discriminator ($\mathcal{D}_{img}$) and In Table~\ref{tab:inst_da}, we show the architecture details for instance-level domain discriminator ($\mathcal{D}_{inst}$).

\begin{table}[h]
\caption{Image-level domain discriminator}
\begin{center}
\begin{tabular}{|c|}
\hline
Gradient Reversal Layer \\ \hline
Conv, 1 × 1, 64, stride 1, ReLU \\ \hline
Conv, 3 × 3, 64, stride 1, ReLU \\ \hline
Conv, 3 × 3, 64, stride 1, ReLU \\ \hline
Conv, 3 × 3, 3, stride 1 \\ \hline
\end{tabular}
\end{center}
\label{tab:img_da}
\end{table}

\vskip -5mm

\begin{table}[h]
\caption{Instance-level domain discriminator}
\begin{center}
\begin{tabular}{|c|}
\hline
Gradient Reversal Layer \\ \hline
FC, 4096, 1024, ReLU, Dropout \\ \hline
FC, 1024, 1024, ReLU, Dropout \\ \hline
FC, 1024, 2 \\ \hline
\end{tabular}
\end{center}
\label{tab:inst_da}
\end{table}

\clearpage


\begin{figure*}
\begin{center}
\includegraphics[width=0.24\linewidth]{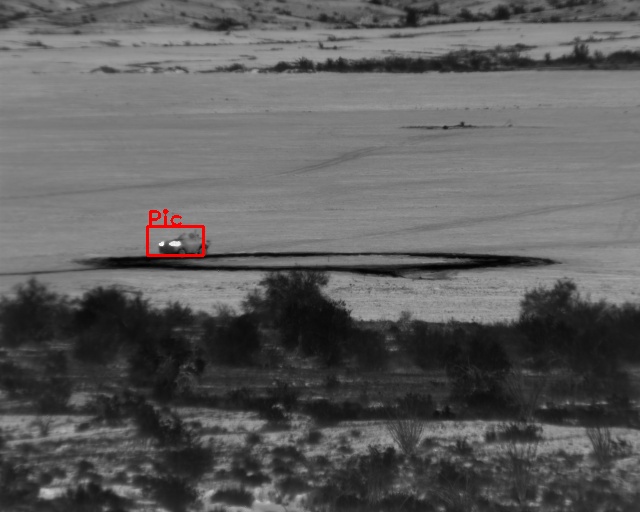}
\includegraphics[width=0.24\linewidth]{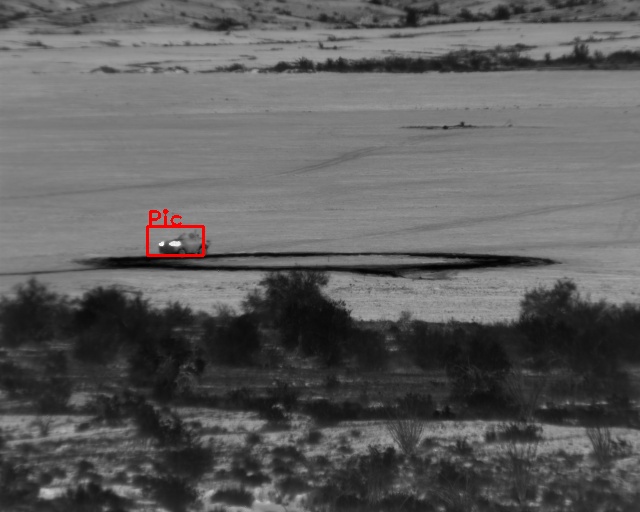}
\includegraphics[width=0.24\linewidth]{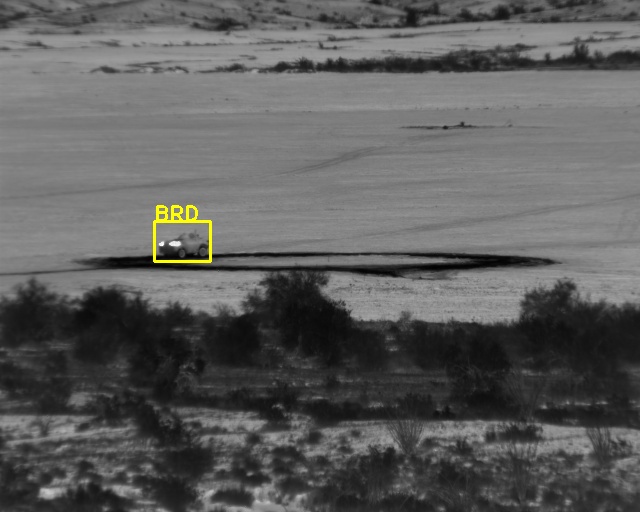}
\includegraphics[width=0.24\linewidth]{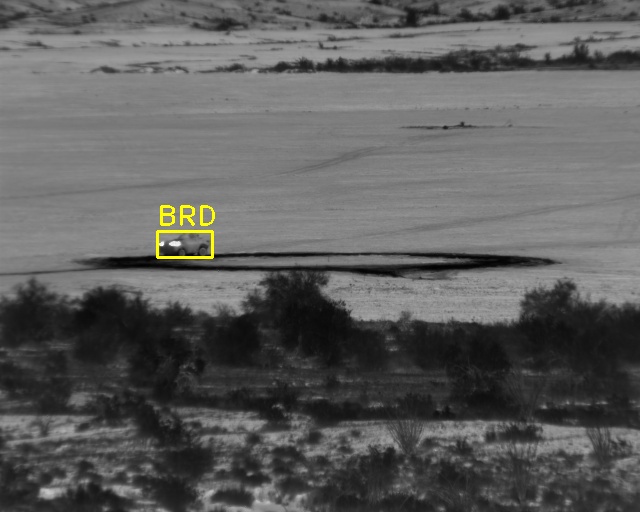}
\includegraphics[width=0.24\linewidth]{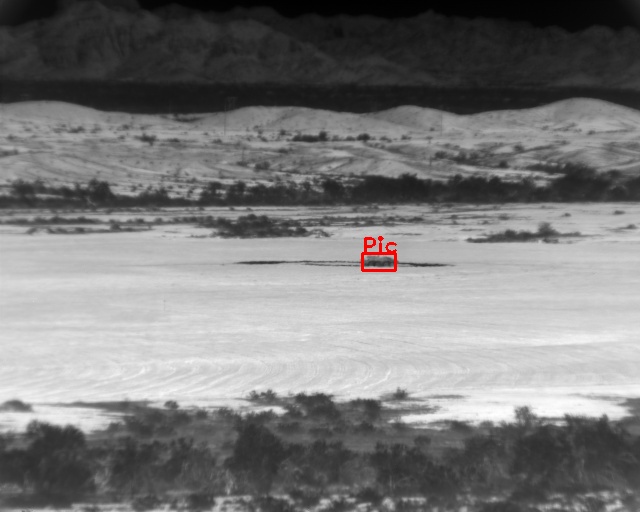}
\includegraphics[width=0.24\linewidth]{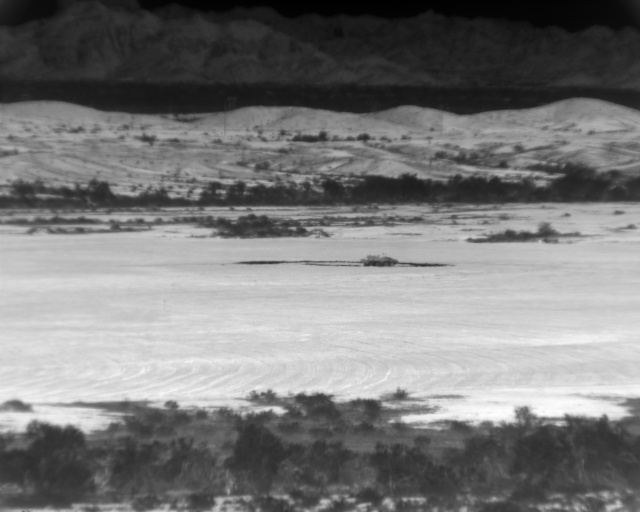}
\includegraphics[width=0.24\linewidth]{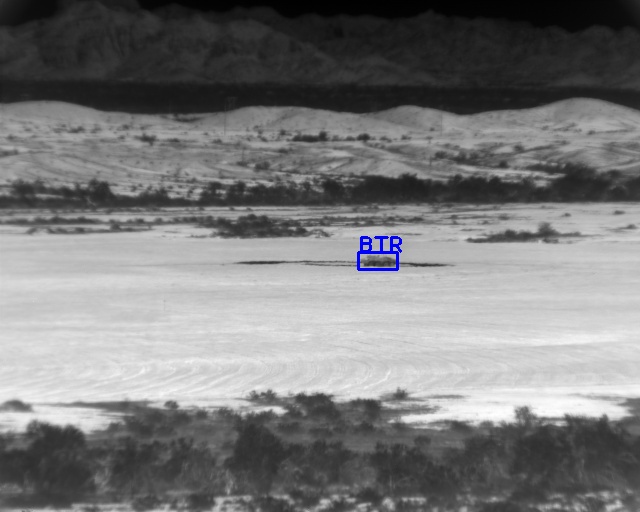}
\includegraphics[width=0.24\linewidth]{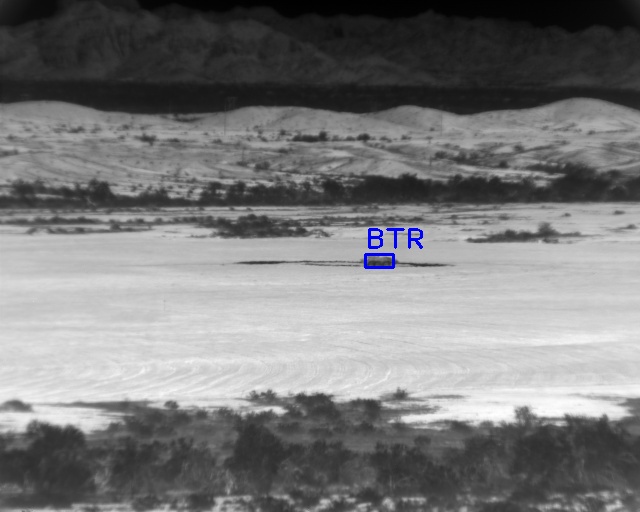}
\includegraphics[width=0.24\linewidth]{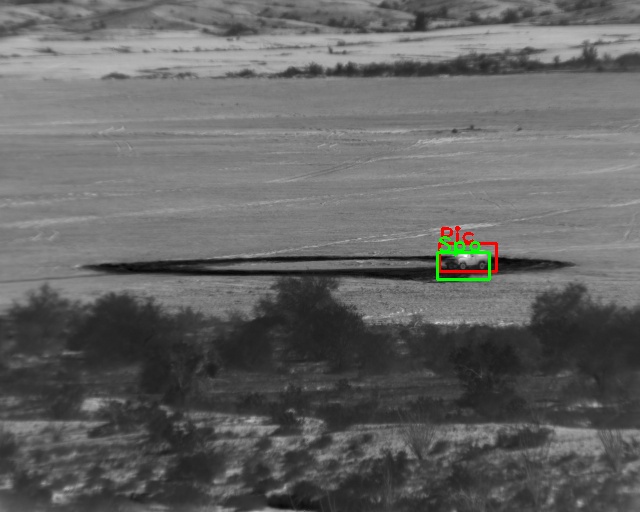}
\includegraphics[width=0.24\linewidth]{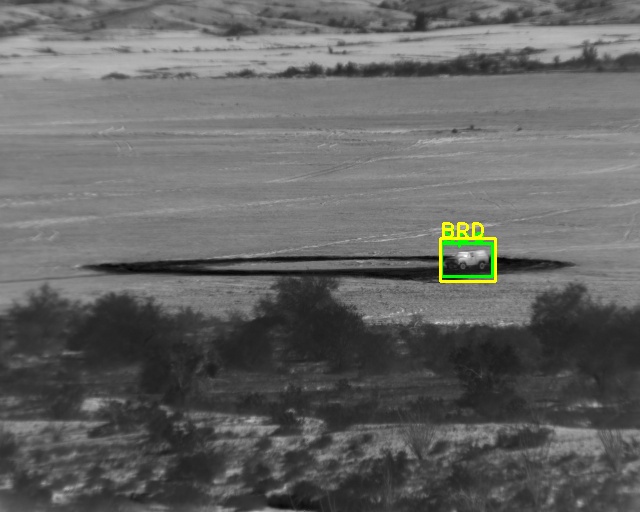}
\includegraphics[width=0.24\linewidth]{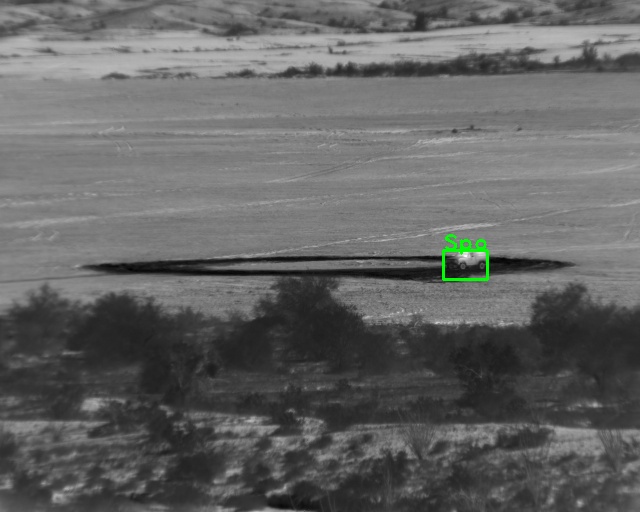}
\includegraphics[width=0.24\linewidth]{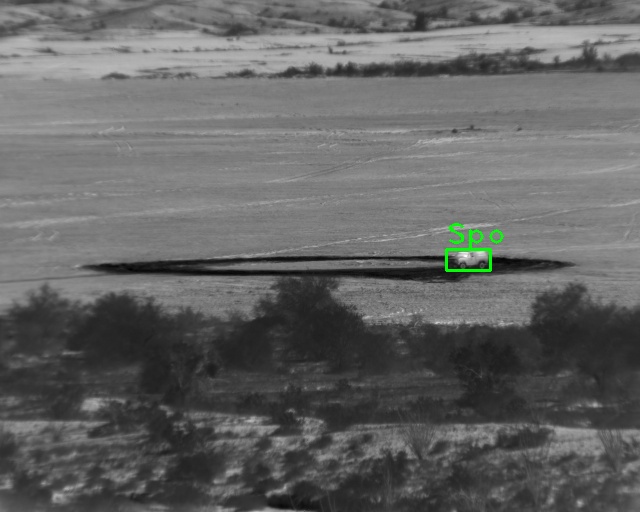}
\includegraphics[width=0.24\linewidth]{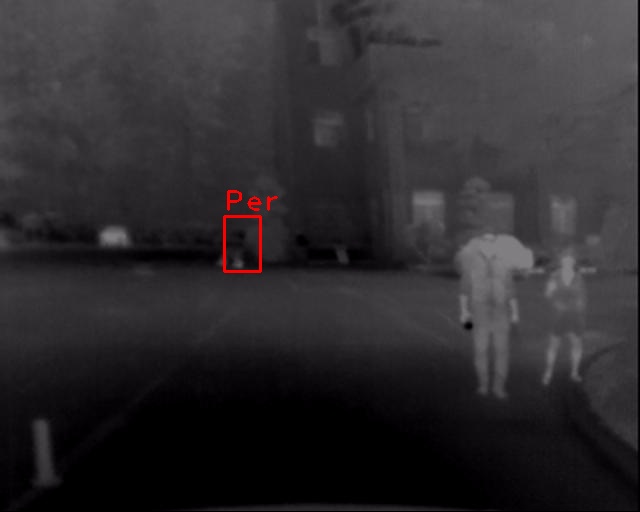}
\includegraphics[width=0.24\linewidth]{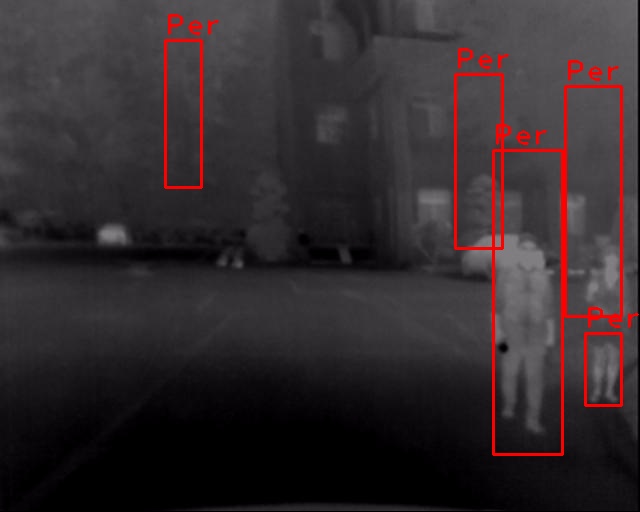}
\includegraphics[width=0.24\linewidth]{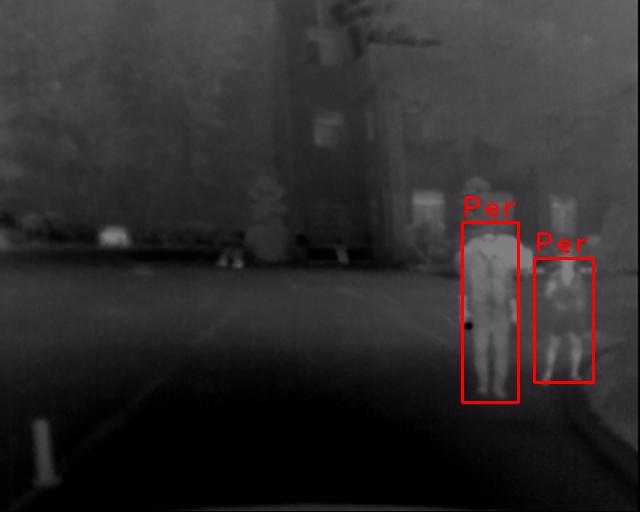}
\includegraphics[width=0.24\linewidth]{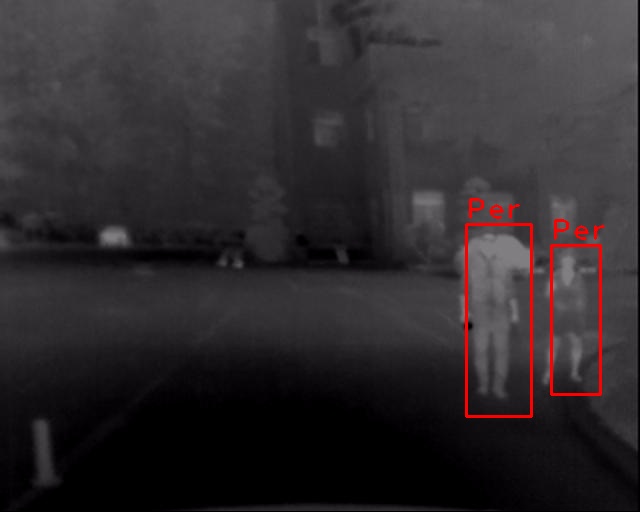}
\includegraphics[width=0.24\linewidth]{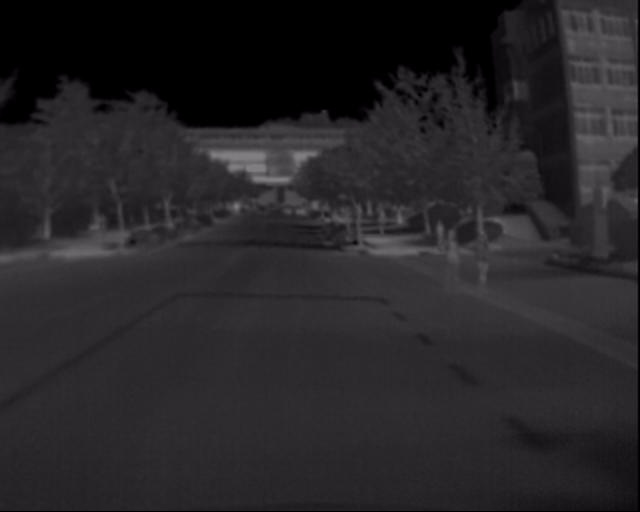}
\includegraphics[width=0.24\linewidth]{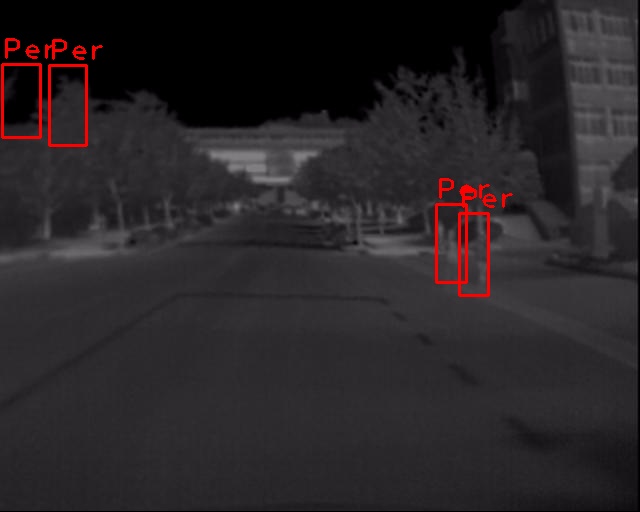}
\includegraphics[width=0.24\linewidth]{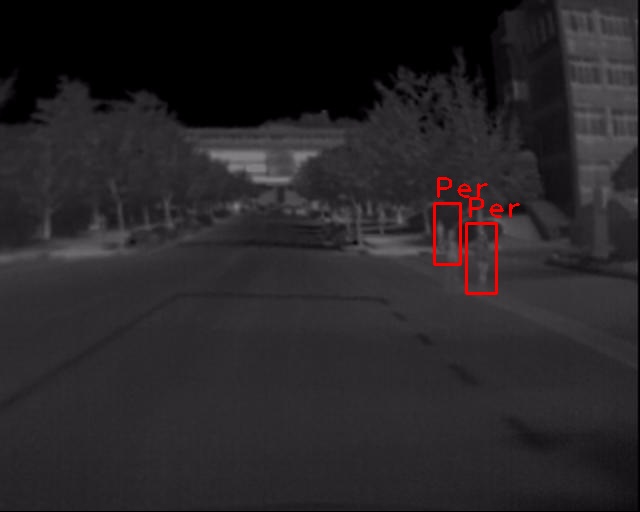}
\includegraphics[width=0.24\linewidth]{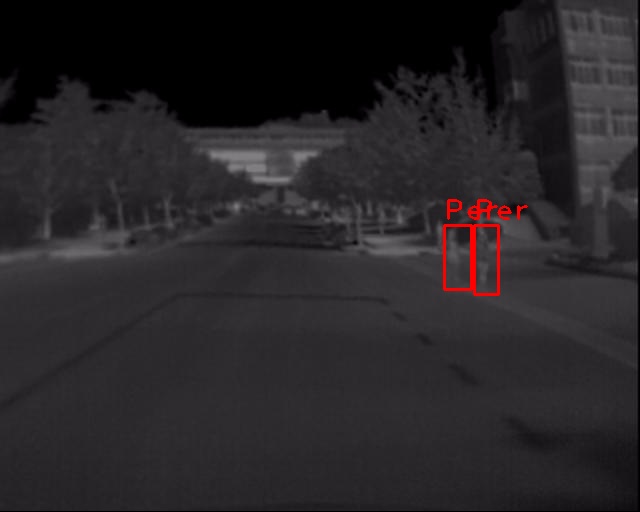}
 Source only \hskip 60pt DA Faster-RCNN \hskip 75pt Ours \hskip 85pt Ground Truth
\end{center}
\vskip -5.0pt \caption{More detection visualization for visible $\rightarrow$ thermal adaptation for DSIAC and KAIST dataset. We show detections with scores higher than 0.5. In the DSIAC dataset, source only and DA Faster-RCNN produce false-positive predictions, whereas our method recognizes the object correctly. Similarly, in the KAIST dataset, our method reduces false-positive as well as produces a high-quality prediction. Because meta-learning helps in achieving fine DA updates resulting in a more robust and generalized detector.}
\label{fig:sup_foggy}
\end{figure*}

\end{document}